\documentclass[letterpaper]{article} 
\usepackage{aaai23}  
\usepackage{times}  
\usepackage{helvet}  
\usepackage{courier}  
\usepackage[hyphens]{url}  
\usepackage{graphicx} 
\usepackage{natbib}  
\usepackage{caption} 
\usepackage{algorithm}
\usepackage[noend]{algorithmic}
\usepackage{arydshln}

\usepackage{newfloat}
\usepackage{listings}
\DeclareCaptionStyle{ruled}{labelfont=normalfont,labelsep=colon,strut=off} 
\floatstyle{ruled}
\newfloat{listing}{tb}{lst}{}
\floatname{listing}{Listing}
\usepackage{color}
\usepackage{amssymb}

\title{Meta-operators for Enabling Parallel Planning \\ Using Deep Reinforcement Learning}

\author {
    Ángel Aso-Mollar\textsuperscript{\rm 1},
    Eva Onaindia\textsuperscript{\rm 1}
}
\affiliations {
    \textsuperscript{\rm 1} Valencian Research Institute for Artificial Intelligence (VRAIN),\\ Universitat Politècnica de València (UPV)\\
    aaso@vrain.upv.es, onaindia@dsic.upv.es
}

\usepackage{bibentry}

\begin{document}

\maketitle

\begin{abstract}
There is a  growing interest in the application of Reinforcement Learning (RL) techniques to AI planning with the aim to come up with general policies. Typically, the mapping of the transition model of AI planning to the state transition system of a Markov Decision Process is established by assuming a one-to-one correspondence of the respective action spaces. In this paper, we introduce the concept of meta-operator as the result of simultaneously applying multiple planning operators, and we show that including meta-operators in the RL action space enables new planning perspectives to be addressed using RL, such as parallel planning. Our research aims to analyze the performance and complexity of including meta-operators in the RL process, concretely in domains where satisfactory outcomes have not been previously achieved using usual generalized planning models. The main objective of this article is thus to pave the way towards a redefinition of the RL action space in a manner that is more closely aligned with the planning perspective.
\end{abstract}

\section{Introduction}

Integrating AI planning and Machine Learning algorithms is a hot research topic that has reported significant advances in many different scenarios, such as training general policies with supervised or unsupervised Learning, \cite{bonet22b}, \cite{bonet22a}, Generalized Planning using Reinforcement Learning (RL), \cite{asai22}, \cite{rivlin20}, \cite{frances19}, learning Action Models using Reinforcement Learning \cite{alvin19}, and many others.

Specifically, strands of this discipline have integrated AI Planning with RL usually to train, with synthesized experience, heuristics that guide the search process of a planning agent, through rectification of domain-independent heuristics \cite{asai22} or through direct policy training \cite{rivlin20}, for example. The training process is usually based on giving the agent a reward of 1 when it reaches the goal of the planning problem and 0 otherwise. This introduces the so-called sparse reward problem wherein we observe that the agent rarely receives rewards from the environment, thus hindering its ability to learn. It is also known that some works attempt to overcome these problems using, for instance, Reward Machines \cite{rewardmachines}. 

In addition, it is common for papers that use RL with AI planning to make use of the very action definition in planning by applying it directly into Reinforcement Learning: to establish the mapping between the transition model of AI planning and the state transition system of a Markov Decision Process, it is assumed that there is a one-to-one correspondence between their respective action spaces. This is what we refer to as the 1-1 mapping between planning and RL actions. In this paper, we propose a new perspective based on the concept of meta-operator, with which we decouple this 1-1 mapping often seen in the literature, therefore allowing to apply several actions at the same time (in parallel) in a concrete time step. Parallel plans increase the efficiency of planning because increased parallelism leads to a reduction of the plan length or the number of time points. In this paper, we also investigate if general parallel policies can be obtained with a generalized planning framework and if such policies improve general sequential policies.

By defining the concept of meta-operator as the application of several planning actions at the same time, we can force the RL training process to simulate parallelism. In addition, we can also use meta-operators to further guide the training process, alleviating the sparse reward problem by establishing a small reward when the meta-operator contains more than one atomic action. 

We incorporate meta-operators in the context of Generalized Planning and to this aim our approach is based on a modified version of the architecture proposed in \cite{rivlin20}, where Graph Neural Networks are used to generate a compact representation of the planning states. We will report results showing that the inclusion of meta-actions allows almost a 100\% coverage in domains where it is difficult to reach generalization with traditional learning methods such as \textit{logistics} or \textit{depots}, as stated in \cite{bonet22b}. 

We tested our models in problem instances from the International Planning Competitions\footnote{\url{https://www.icaps-conference.org/competitions/}} (IPC) and in randomly generated ones, and we observed that the coverage using meta-operators improves with respect to not using them. It is also observed that the length of the plan, as expected, is also reduced when we include meta-operators.

This paper is structured as follows: several basic concepts with which we rely our work on are going to be defined, such as Planning, RL and their integration, including Generalized Planning. Next, the core element of this article will be defined: the meta-operator, and then will be integrated into the RL structure. Finally, several different experiments that justify the use of meta-operators in RL will be conducted, followed by a discussion of whether to use this mechanism and the benefits against other models that do not include this approach.

\section{Background}

\subsection{Planning}

Classical planning is the problem of finding a sequence of actions that when applied to an initial state lead to the achievement of a goal or set of goals. The two main components of a planning task are the domain and the problem.

A \textbf{planning domain} is defined as $D=\langle\mathcal{F},\mathcal{A}\rangle$, where $\mathcal{F}$ is a set of fluents that describe lifted properties of the domain objects and their relations, and $\mathcal{A}$ is a set of action schemas in which every $a \in \mathcal{A}$ is defined by a triplet $\langle\textup{Pre}(a),\textup{Add}(a),\textup{Del}(a)\rangle$,\ $\textup{Pre}(a), \textup{Add}(a), \textup{Del}(a) \subseteq \mathcal{F}$.

A \textbf{planning problem instance} linked to a domain $D=\langle\mathcal{F},\mathcal{A}\rangle$ is defined as a tuple $P = \langle F,O,I,G \rangle$ where $F$ is the set of \textit{facts} that result from grounding the fluents $\mathcal{F}$ to the objects of the problem $P$, and $O$, called \textit{operators}, is the result of grounding the action schemas $\mathcal{A}$. A state $s \subseteq F$ is a set of facts, and $I,G \subseteq F$ are, respectively, the initial and the goal state of the problem. 

Every operator $o \in O$ has its corresponding grounded triplet defined by $\langle Pre(o),Add(o),Del(o)\rangle$. $Pre(o) \subseteq F$ are the preconditions that must be true in a state for the operator to be applicable; that is, an operator $o$ is \textit{applicable} in a state $s \subseteq F$ if $Pre(o) \subseteq s$. $Add(o) \subseteq F$ are the effects of the operator that assert a positive literal after the application of $o$; and $Del(o) \subseteq F$ is the set of negative effects, i.e., the set of literals which become false after the operator is applied. Applying $o \in O$ in $s$ leads to a new state $s'=(s \setminus Del(o)) \cup Add(o))$.

A solution to a planning problem $P = \langle F,O,I,G \rangle$ is a sequence of operators or \textit{plan} $\rho= \langle o_1,o_{2},\ldots,o_{k}\rangle, \ o_{i} \in O\ \forall i$, such that the result of applying the operator sequence $\langle o_{1},o_{2},\ldots,o_{k}\rangle$ to the initial state $I$ leads to a state $s\subseteq F$ that satisfies $G \subseteq s$, that is to say, that the state $s$ contains every goal in $G$. 

\subsection{Reinforcement Learning}

Reinforcement Learning (RL) is a computational approach to learning from environmental interaction \cite{sutton98}. The main objective of RL is to learn a \textit{policy}, or behaviour, that maximizes the positive interaction that an agent gets through time.

RL scenarios are often modeled as finite Markov Decision Processes (MDP) \cite{mdpsbasic}. An MDP is a control process that stochastically models decision-making scenarios; an agent continuously interacts with an environment by executing actions that change its internal state, and these actions are rewarded consequently. The main objective of the agent is to improve its performance so that it can progressively maximize the received reward.

Formally, an MDP is defined as $M=\langle S,A,R,Pr\rangle$,
where $S$ is a set of states, $A$ is a set of actions, $R : S \times A \times S \rightarrow \mathbb{R}$ is a reward function that values how good or bad it is to take an action $a_t$ at a certain state $s_t$ that leads to another state $s_{t+1}$, $R(s_t,a_t,s_{t+1}) = r_t$, and $Pr$ is a transition probability function such as $Pr(s',s,a)=p(s'|s,a)$, which is usually unknown. 

At each time step $t$, an agent takes an action in the current state $s_t$ among a set of applicable actions $A_t \subseteq A$ in $s_t$ following a probability distribution called \textit{policy}, represented by $\pi: S \rightarrow p(\cdot |\ S = s_t)$. The objective of RL is to reward decisions that reach the target as quickly as possible, by prioritizing immediate rewards. 

The RL optimization problem is formally defined as 
\begin{equation} 
\label{optmization}
    \max_{\pi} \mathbb{E}_\pi G_t
\end{equation}

where $G_t = \sum_{k=0}^{\infty}\gamma^kr_{t+k+1}$ is the sum of the discounted reward of a trajectory sampled from the policy $\pi$ from the current state $t$ to the infinite. Thus, formula (\ref{optmization}) measures the reward received by all possible trajectories multiplied by the probability of taking the decisions of the trajectories under the policy $\pi$, where $\gamma \in [0,1]$ is the discount factor of future rewards.

\subsection{Integrating RL on top of planning}

The integration of Planning and RL is a promising approach for solving control problems. Planning is effective to reason over a long horizon but assumes access to a local schema; in contrast, RL is effective at learning policies and relative values of states but fails to plan over long horizons \cite{EysenbachSL19}. Combining planning and RL can thus be regarded as learning a control policy for the planner in lieu of using a planning heuristic function. 

Combining Planning and RL enables addressing a variety of tasks like learning generalist policies to solve a large number of problems \cite{celorrio19}, learning action models based on interaction \cite{modelbasedrlsurvey}, training intelligent heuristics that guide the agent in the control task \cite{asai22}, or even training the agent to solve concrete planning problems. Generally, the RL module is integrated \textit{inside} the planning agent to help it make a decision, although other approaches use it the opposite way \cite{hierarchicalRL}. In this work, we will follow the first approach.

As the formulation of both RL and Planning use the concept of state and action, we must establish the equivalence between RL actions and planning operators, so that the planning and RL modules can communicate with each other. 

It is usually assumed in the literature that for a planning problem $P=\langle F,O,I,G\rangle$  the equivalence is established by simply considering the planning operators $O$ as the RL actions $A$ ($A=O$).
States in RL ($S$) are abstract concepts that simply represent situations that occur in the environment, so we can inherit the concept of planning state and apply it directly to RL by making every $s \in S$ as $s \subseteq F$; i.e., we can assume RL states as planning states. 
Hence, the notion of an action being applicable to a state $s$ in RL follows directly from the above considerations and uses the same principle of applicability of an operator in the planning state $s$. 

Since this one-to-one correspondence between actions and operators yields huge action spaces, some researchers adopt certain considerations to reduce the number of possible actions. For example, in PDDLGym \cite{pddlgym}, authors consider only a partial grounding of the action schema in order to reduce the number of possible actions. Particularly, they claim that certain operators contain parameters that correspond to \textit{free} objects (in terms of controlling the agent), that is, objects whose properties can potentially change, leaving ungrounded the \textit{non-free} agents. For example, in a transport domain in which we want to transport packages between cities using trucks, the packages and trucks would be free objects because their location changes, but the cities would not. In this scenario, an action schema like \texttt{(move ?t - truck ?c1 - city ?c2 - city)} will only bound the parameter \texttt{truck} to a concrete object as the two parameters \texttt{city} refer to static objects that only denote the structure of the movement network.

We can think of this consideration as a transformation $g$ of the planning operator space, $g(O)$, which is then used as the RL action space, $A=g(O)$. Note that in PDDLGym this specification is hard-coded, while we propose an automated method, as we will discuss later.

We already mentioned that the notion of a planning state is directly translated into an RL state. We need to take into account that sometimes, and depending on the RL method, the planning states must be translated into some encoding that the RL method supports. For example, if we are using a NN to represent the policy, we need a vectorial representation of the state. This representation can be obtained, for example, by using Graph Neural Networks \cite{gnnsurvey} or Neural Logic Machines \cite{nlmpaper}, in conjunction with intermediate structures such as graphs. 

Sometimes we will say that we \textit{apply} a learned policy $\pi$ to a planning problem $P$, meaning that the planning operator corresponding to the RL action given by the policy is successively applied from the initial state of the problem to the goal, and states will change accordingly through the planning operators application. This way, as mentioned above, the planning states are mapped to the RL states and (hegemonically) planning operators are mapped to RL actions.  

\subsection{Generalized Planning}

Generalized Planning is a discipline in which we aim to find general policies that explain a set of problems, usually of different size. Specifically, given a set of problems $\{\langle F_i,O_i,I_i,G_i\rangle\}_{i=1}^N, N>0$, rather than searching for a solution plan $\rho_i = \langle o_{1},o_{2},\ldots,o_{k}\rangle,\ o_{j} \in O_i$ for each individual problem $P_i$ by applying a policy $\pi_i$ previously trained for $P_i$, we aim to find a general policy $\pi$ that when applied to every problem of the set returns a solution plan to all of them.

\section{Defining meta-operators}

As we discussed in the previous section, a bijective correspondence is usually considered between the set of planning operators and the set of RL actions when solving a planning problem $P=\langle F,O,I,G \rangle$ of a domain $D$. That is, $A = O$. We have also seen that we can transform this operator space into a different one $g(O)$. In this work, unlike PDDLGym that performs a reduction of the operator space, we will define a transformation function to enrich it.

For this purpose, we define the notion of a \textbf{meta-operator} for a problem $P=\langle F,O,I,G \rangle$ as a new operator that, given $L > 1$ and $o_1,...,o_L \in O$, which we will refer to as \textit{atomic operators}, is represented as
$$\bigoplus_{i=1}^L o_i$$
with the following considerations:

\begin{itemize}
    \item $Pre\left(\bigoplus_{i=1}^L o_i\right) = \bigcup_{i=1}^L Pre(o_i)$
    \item $Add\left(\bigoplus_{i=1}^L o_i\right) = \bigcup_{i=1}^L Add(o_i)$
    \item $Del\left(\bigoplus_{i=1}^L o_i\right) = \bigcup_{i=1}^L Del(o_i)$
    \item No pair of actions $o_i$, $o_j$ from all the atomic actions $o_1,...,o_L$ that form the meta-operator \textit{conflict} with each other. Two atomic operators $o_i$ and $o_j$ \textbf{conflict} with each other if one of these two conditions hold:
\begin{itemize}
    \item $\exists\ p \in Pre(o_i)$ such that $p \in Del(o_j)$.
    \item $\exists\ p \in Add(o_i)$ such that $p \in Del(o_j)$.
\end{itemize}
\end{itemize}

Broadly speaking, a meta-operator is nothing but a synthetic operator resulting from the union of atomic ones that can be executed in any order. The resulting operator will therefore inherit the union of its sets $Add$, $Pre$ and $Del$, always bearing in mind that all the atomic operators involved can be executed at the same time, that is, that they do not \textit{conflict} with each other.

This means that they are inconsistent if they compromise the consistency of the resulting state when applying the operator, i.e., if the resulting state changes if the sequential order of application of the atomic operators changes. These two considerations are equivalent to the notions of \textit{inconsistent effects} and \textit{interference} in the calculation of a mutex realtion between two actions in GraphPlan \cite{BlumF97}.

We define the set of meta-operators of degree $L \in \mathbb{N}$ in a problem $P$ of a domain $D$ as the union of every possible meta-operation:
$$\mathcal{O}^L = \bigcup_{o_i \in O} \left[\bigoplus_{i=1}^L o_i \right]$$
where $o_i$ are actions from $O$ that do not interfere with each other when defining a single meta-operator and $\mathcal{O}^1=O$.

\section{Including meta-operators in RL}

Meta-operators are then added to the RL action space, enriching and enabling the application of parallel planning operators within this sequential RL action space. We can therefore define a new transforming function $g_L$ as the union of the base set with meta-operators of degree $L$:
$$g_L(O) = \bigcup_{i=1}^L \mathcal{O}^i$$
and train our RL algorithms using this consideration, i.e., $A = g_L(O)$. 

This integration of meta-operators is calculated online, as in Algorithm \ref{alg:algorithmappl}, at each time step $t$ and current state $s_t$, for all applicable RL actions $A_t = \{o \in O : o \textup{ is applicable at } s_t\}$. We define that a meta-operator is \textit{applicable} at state $s_t$ if every atomic operator is also applicable at $s_t$ and if operators do not interfere with each other, which is followed by definition.

\begin{algorithm}[tb]
\caption{Calculate all applicable RL actions}
\label{alg:algorithmappl}
\textbf{Input}: \\
A set of applicable planning operators $O$\\ Degree of meta-operators $L$ \\
\textbf{Output}: \\ A set of applicable RL actions $A$

\begin{algorithmic}[1] 
\STATE $A = O$
\STATE $N = \emptyset$ 

\STATE \# Generate all conflicts

\FOR{$a \in O$}

\FOR{$p \in Pre(a)$}
\FOR{$b \in O$}
\IF{$p \in Del(b)$}
\STATE $N = N \cup \{(a,b),(b,a)\}$ \# Interference
\ENDIF
\ENDFOR
\ENDFOR

\FOR{$p \in Add(a)$}
\FOR{$b \in O$}
\IF{$p \in Del(b)$}
\STATE $N = N \cup \{(a,b),(b,a)\}$ \# Inconsistent effects
\ENDIF
\ENDFOR
\ENDFOR
\ENDFOR

\STATE \# Combine pairs, triplets, until $L$, of operators
\STATE \# that do not contain any pair of operators in $N$.
\FOR{$i \in \{2,...,L\}$}
\STATE $ A = A \cup \textup{MakeMetaOperators}(O,L,N)$
\ENDFOR

\STATE \textbf{return} A
\end{algorithmic}
\end{algorithm}

We used the Generalized Planning RL training scheme proposed in \cite{rivlin20} to observe the effects of the inclusion of a meta-operator in domains that often struggle to generalize in the literature, such as \textit{logistics} or \textit{depots} \cite{bonet22b}. That architecture also uses GNNs for state representation and gives a reward of 1 to the agent if it reaches the goal, and 0 in any other case, as usual.

This last decision is highly criticized by the planning community because it introduces the so-called sparse rewards problem, that is, the agent receives information from the environment at very specific moments, thus hindering the learning process. The inclusion of meta-operators opens up the possibility of defining a certain reward for actions that include more than one atomic operator, briefly alleviating the aforementioned problem.

In that sense, we want to test how this inclusion of different specific amounts of reward in meta-operators affects to the learning process, so we will do an analysis using different amounts of reward and we will see how well they train and how parallel generated plans are compared with each other.

We also want to test whether the inclusion of meta-operators actually improves the coverage for problems of domains we analyzed, compared to the coverage of a sequential model, trained with the same domain but without using meta-operators ($A=O$).

\section{Experiments}

In this section, we present a series of experiments that support the inclusion of meta-operators in Generalized Planning using RL. In particular, we are interested in two things: (1) analyzing the impact of an extra reward when a meta-operator is applied in the learning process, and (2) checking whether the inclusion of meta-operators improves the results in terms of coverage (number of solved problems).

Specifically, we conduct two experiments. Experiment 1 is designed to measure the degree of parallelism of the solution plans using different rewarding in meta-operators. Experiment 2 evaluates the performance of our model against two different defined datasets.

\subsection{Domains}

We will use two domains that are widely used in the IPC and also known for their complexity, \textit{logistics} and \textit{depots}, and a third domain which is an extension of the well-known \textit{blocksworld} domain.

\subsubsection{Multi-blocksworld.} This domain is an extension of the \textit{blocksworld} domain that features a set of blocks on an infinite table arranged in towers, with the objective of getting a different block configuration by moving the blocks with robot arms. Blocks can be put on top of another block or on the table, and they can be grabbed from the table or from another block. We have defined two robot arms.

\subsubsection{Logistics.}This domain features packages located at certain points which must be transported to other locations by land or air. Ground transportation uses trucks and can only happen between two locations that are within the same city, while air transportation is between airports, which are special locations of the cities. The destination of a package is either a location within the same city or in a different city. In general, ground transportation is required to take a package to the city's airport (if the package is not at the airport). The package is then carried by air between cities, and finally using ground transportation  the package is delivered to the final destination if its destination is not the arrival airport. 

\subsubsection{Depots.}This domain consists of trucks that are used for transporting crates between distributors, and hoists to handle the crates in pallets. Hoists are only available at certain locations and  are static. Crates can be stacked/unstacked onto a fixed set of pallets. Hoists do not store crates in any particular order. This domain slightly resembles the \textit{multi-blocksword} domain as there is a stacking operation, though crates do not need to be piled up in a specific order, and to the \textit{logistics} domain as to the existence of agents that transport crates from one point to another.

\subsection{Data generation}

RL algorithms need a large number of instances in order to converge. That is why, for the training process, it was necessary to use automatic generators of planning problems. For \textit{logistics} and \textit{depots} domains, we used generators of the AI Planning Github \cite{pddlgenerators}, while for the \textit{multi-blocksworld} domain we created a new generator based on the generator for the \textit{blocksworld} domain \cite{pddlgenerators}.

\begin{table*}
    \centering
    \begin{tabular}{c c c c c}
        \hline
        \hline
        \textbf{Domain} & \textbf{Train size}  &  \textbf{Total objects train} &
        \textbf{Validation/Test size} &
        \textbf{Total objects test} \\
        \hline
        \hline
        Multi-blocksworld &
        5-6 blocks & 5-6 & 10-11 blocks &
        10-100 \\
        \hline
        Logistics &
        \begin{tabular}{c}2-4 airplanes \\ 2-4 cities \\ 2-4 trucks \\ 2-4 locations per city \\ 1-3 packages\end{tabular} 
         & 9-10 &
        \begin{tabular}{c}3-4 airplanes \\ 6-7 cities \\ 3-4 trucks \\ 6-7 locations per city \\ 6-7 packages\end{tabular} & 24-29 
       \\
        \hline
        Depots &
        \begin{tabular}{c}1-2 depots \\ 2-3 distributors \\ 2-3 trucks \\ 3-5 pallets \\ 2-4 hoists \\ 3-5 crates\end{tabular} 
         &  13-22 & 
        \begin{tabular}{c}5-6 depots \\ 5-6 distributors \\ 5-6 trucks \\ 5-6 pallets \\ 5-6 hoists \\ 5-6 crates\end{tabular}  & 30-36
         \\
        \hline
        \hline
   \end{tabular}
    \caption{Sizes used for the problem generation, in terms of general and specific objects.}
    \label{tab:trainsizes}
\end{table*}

Table \ref{tab:trainsizes} illustrates the size distribution of the problems used in this work; it shows the number of objects of each type involved in all three domains. We generated a dataset of random problems out of the distributions shown in Table \ref{tab:trainsizes}, which we will refer to as \textbf{Dataset 1 (D1)}.

\subsubsection{Dataset 1 (D1)} It consists of problems uniformly sampled from the test distribution of Table \ref{tab:trainsizes} and generated with the aforementioned generators. We generated 460 problems for the \textit{multi-blocksworld} domain, 792 problems for the \textit{logistics} domain and 640 for the \textit{depots} domain, as a result of creating ten problems for each configuration in the test distribution.

\vspace{0.1cm}

Additionally, we created a second collection of samples that we will refer to as \textbf{Dataset 2 (D2)} from a renowed planning competition.

\subsubsection{Dataset 2 (D2)} It consists of problems that were used in the IPC (specifically, in the IPC-2000 and IPC-2002). We used the 35 first instances for the \textit{blocksworld} domain in IPC-2000, with a slight modification to introduce the two robot arms; the 30 first instances of \textit{logistics} from IPC-2000; and the 22 instances of \textit{depots} from IPC-2002. This set of instances was chosen in order to compare our results with those obtained in the work \cite{bonet22a}.

\subsection{Setup}

We opted for using a $L=2$ degree meta-operator to come up with a feasible extension of the action space. As the inclusion of meta-operators increases the action space, we need to find a balance between size and performance. Using two-degree meta-operators is sufficient to fulfill the two objectives mentioned at the beginning of this section, namely analyzing the impact of rewarding meta-operators and evaluating the coverage of the models. We will also evaluate how much does the action space rise with our approach compared to a sequential model.

The RL training was conducted on a machine with a Nvidia GeForce RTX 3090 GPU, a 12th Gen Intel(R) Core(TM) i9-12900KF CPU and Ubuntu 22.04 LTS operating system, and the same hyperparameter configuration than \cite{rivlin20}. A similar training process as the one proposed in \cite{rivlin20} was followed here, all policies are trained for 900 iterations, each with 100 episodes and up to 20 gradient update steps, using Proximal Policy Optimization RL training algorithm, with a discount factor of 0.99.

\subsection{Experiment 1: Rewarding of meta-operators}

In this experiment, we aim to observe how the reward granted to the application of a meta-operator in the RL  training influences the learning process and the quality of the plans. We are interested in measuring the effect of meta-operators in terms of the plan length or the number of time steps of the plan. To that end, we define the \textit{parallelism rate} of a solution plan of a problem as:
$$\frac{\#\ parallel\ operators}{\#\ total\ plan\ timesteps}$$

where $\#\ parallel\ operators$ is the number of meta-operators that appear in the plan by applying the learned policy to the problem, and $\#\ total\ plan\ timesteps$ is the total number of timesteps of the plan. This is a measure of how frequently parallel operators appear in the decisions made by the planner agent.

We trained a series of models giving different reward values to meta-operators. This experiment can be thought of as a way of tuning the meta-operators reward, which can therefore be regarded as a hyperparameter. Since we primarily aim to find the most appropriate reward for the use of meta-operators in this experiment, we decided to focus only on the training distribution.

We trained five models from the train distribution of Table \ref{tab:trainsizes} with reward values to meta-operators of $0.0, 0.1, 0.01, 0.001$ and $0.0001$, respectively. Subsequently, the five models were run on a fixed sample, generating 10 problems for each element from the train distribution, and results were analyzed in order to obtain the \textbf{average} parallelism rate of all plans. 

During the experiment execution, rewards and the number of parallel actions at each time step are monitored so as to balance out the reward coming from parallel actions and coming from achieving a solution plan. In other words, we want to avoid situations in which parallel actions are just added for the sake of reward, which may deviate plans towards a large number of parallel actions sacrificing reaching the objective.

\begin{table}[t]
    \centering
    \begin{tabular}{c c c c}
    \hline
    \hline
         \textbf{Reward} & \textbf{Multi-blocks} & \textbf{Logistics} & \textbf{Depots} \\
         \hline
         \hline
         0.0 & 0.550 & 0.243 & 0.530\\
         \hline
         0.1 & - & - & - \\
         \hline
         0.01 & 0.701 & 0.851 & 0.857\\
         \hline
         0.001 & 0.559 & 0.582 & 0.381\\
         \hline
         0.0001 & 0.557 & 0.768 & 0.294\\
         \hline
         \hline
    \end{tabular}
    \caption{Average parallelism rate for all models trained with specified reward for the application of meta-operators.}
    \label{tab:paralrate}
\end{table}

The results of this experiment are shown in Table \ref{tab:paralrate}: all models are able to fulfill the aforementioned objective (100\% of coverage in training) except for the model that gives a reward of 0.1 to meta-operators (no results are shown because the model did not converge). Intuition tells us that there are certain values that reward too much parallelism, even above reaching the problem goal itself, resulting in potentially infinite plans that execute parallel actions in a loop (until the maximum episode time is reached). 
This means that a reward value of 0.1 for meta-operators outputs policies that yield more than ten parallel actions per plan,  which exceeds the value given to reaching the goal, thus making the algorithm converge to a situation in which no goal is reached but lots of meta-operators of degree $L > 2$ appear in the plan. Ultimately, RL is about optimizing a reward function, and if adding meta-operators produces more reward, this will be the path taken by the model.

According to Table \ref{tab:paralrate}, we observe that the model that gives the best results in terms of parallelism rate for all domains within the plans generated is the 0.01 reward model. This indicates that, in order to obtain potentially better results, a balance must be established between the amount of reward given to parallelism and the amount of reward given to the goal. 

For example, a somewhat more conservative proposal, which we know for sure would not exceed the goal reward, is to establish a meta-operator reward of $\frac{GOAL\_REWARD}{MAX\_ITERATIONS}$, where $GOAL\_REWARD$ is the reward given to reaching the goal (generally 1) and $MAX\_ITERATIONS$ is the maximum number of applications of the policy before stating that the goal cannot be achieved. Generally, this approach is excessively limiting and does not encourage parallelism. This is evident from Table \ref{tab:paralrate}, which shows that greater rewards lead to improved parallelism.

In fact, the amount of reward provided in meta-operators is also dependent on the average length of the plans we want to test. That means, perhaps if the problems we want to test have a larger average plan length than the ones we trained, it would be wiser to test with a model that has been trained with a slightly lower reward in order to not ``overflow'' the reward and fall into the undesired scenario of policies that produce parallel actions with no goal termination. This would be a problem that would probably occur in Generalized Planning, for example, where we train models with a problem size smaller than the problem size on which we will test the results.

All in all, it has been found that the amount of reward given to meta-operators is significant in terms of quality and convergence of plans.

\subsection{Experiment 2: Performance in Generalized Planning}

In this experiment, we compare the original sequential model with versions of the parallel model obtained with different reward values. We note that the aim is to test the performance of the models when dealing with new inputs of a larger size. The trained policy for each domain is then analyzed as in the Generalized Planning literature by testing the problems in datasets D1 and D2. Particularly, for each model, we measure the coverage and the average length of the generated plans for the problems in D1 and D2.

\begin{table*}
\centering
\begin{tabular}{c c c c c}
\hline
\hline
\textbf{Domain - D1} & \textbf{OR} & \textbf{R = 0.0} & \textbf{R = 0.001} & \textbf{R = 0.01}\\
\hline
\hline
Multi-blocks (460) & 268 & 439 & 408  & 406 \\
\hline
Logistics (792) & 131 & 701 & 717 & 317  \\ 
\hline 
Depots (640) & 287 & 572  & 640 & 552\\
\hline
\hdashline
\hline
Multi-blocks & 79.99 & 76.43 & 65.92 & 92.84\\
\hline
Logistics & 200.50 & 112.61 & 109.05 & 110.25 \\ 
\hline 
Depots &  338.49 & 106.68  & 124.46  & 121.80\\
\hline
\hline
\hline
\hline

\textbf{Domain - D2} & \textbf{OR} & 
\textbf{R = 0.0}  & \textbf{R = 0.001}& \textbf{R = 0.01}\\
\hline
\hline
Multi-blocks (35) & 2 & 35 & 34 & 34 \\
\hline
Logistics (30) & 11 & 26 & 28& 27 \\ 
\hline 
Depots (22) & 20 & 17 & 20 & 20 \\
\hline
\hdashline
\hline
Multi-blocks & 172.00 & 43.17 & 44.59 & 54.12 \\
\hline
Logistics & 136.64 & 115.92 & 120.64 & 121.00 \\ 
\hline 
Depots & 127.45 & 83.59  & 110.20 &  114.25\\
\hline
\hline

\end{tabular}

\caption{Results for datasets \textbf{D1} and \textbf{D2}. The top part of the tables shows coverage out of the total number of instances shown between parenthesis. The bottom part of the tables indicates the average plan length. \textbf{OR} is the original sequential model in \cite{rivlin20}; \textbf{R=0.0} is the model with no meta-operator reward; \textbf{R=0.1} is the model with reward of $0.01$, and \textbf{R=0.001} is the model with reward of $0.001$.}
\label{testresults}
\end{table*}

Table \ref{testresults} shows the results obtained with the sequential \textbf{(OR)} model \cite{rivlin20}, the parallel model trained with no reward \textbf{(R=0.0)}, with a reward of 0.01 on meta-operators \textbf{(R=0.01)} and with a reward of 0.001 on meta-operators \textbf{(R=0.001)} for the International Planning Competition \textbf{(D2)} and randomly generated \textbf{(D1)} datasets. With these experiments we aim to illustrate how the results vary from one model to another depending on the reward, as stated in the previous section. 

The table is divided in two halves, one for each set of problems. In the top part of each half we show results for coverage of the analyzed models with respect to the problems of each set, while in the bottom part of each half the average length of plans for each problem of the set is analyzed. In the top part of each half we show within parentheses the total number of problems that compound that set, and then in each column the number of those for which the model under analysis has managed to reach the target. The bottom part of each half corresponds to the number of time steps with which the models managed to solve each set of problems. We present the average number of actions taking into account only the solved problems.

Results of Table \ref{testresults} show that coverage from the models that use meta-operators improves with respect to the coverage of the sequential model. Moreover, it is observed that for the \textit{logistics} domain, which has been found in the literature to be hard to address for RL approaches \cite{bonet22a}, a very good result is obtained with respect to that of the sequential model. 

Results are correlated to the enrichment of the action space. Table \ref{tab:actions} shows the size of the action space visited during training of the sequential model and the parallel ones. There is only one result in Parallel because the action space is the same for all three models, only the amount of reward given to the meta-operator application varies. We observe that in the \textit{multi-blocks} domain where there are only two agents (two robot-arms), the increase in the number of actions is much less significant than in \textit{logistics} or \textit{depots} domain, in which many more agents (trucks, planes, etc.) coexist at the same time.

Plans also tend to be shorter in terms of average actions when applying meta-operators. In Table \ref{testresults}, results for \textbf{R=0.0} tend to give short plan lengths, which makes sense because as a sparse reward problem plans tend to reach the goal faster thanks to the discount factor and reward only coming from reaching the goal. This sometimes even result in better coverage, such as in \textit{multi-blocks}, while in \textbf{R=0.001} and \textbf{R=0.01} the reward also comes from the meta-operators and not only from reaching the goal.

\begin{table}
    \centering
    \begin{tabular}{c c c c}
    \hline
    \hline
        & \textbf{Multi-blocks} & \textbf{Logistics} & \textbf{Depots} \\
         \hline
         \hline
         \textbf{Sequential} & 100 & 108 & 228 \\
         \hline
         \textbf{Parallel} & 1140 & 3960 & 8519\\
         \hline
         \hline
    \end{tabular}
    \caption{Action space or number of RL actions (planning operators and, when makes sense, meta-operators) visited during training.}
    \label{tab:actions}
\end{table}

The use of a model with lower reward \textbf{R=0.001} has reported significantly better coverage results than the model with \textbf{R=0.01}. In the Generalized Planning task there is a variance in the size of problems tested, resulting also in a variance in the length of its correspondent plans. As the model \textbf{R=0.01} gives a high reward to parallelism, if the size of the plans is too large, parallelism is being rewarded too much. For example, 11 meta-operators would already mask the objective's reward, which is 1, i.e., $11 \cdot 0.01 = 1.1 > 1$.

 \section{Discussion}

 Although there is not yet a standard neural network architecture that suits perfectly the planning frame, the approaches that leverage deep learning techniques that automatically extract structure from high-level data are mostly based on graph convolution to learn state embeddings like Graph Neural Networks \cite{rivlin20,bonet22a} or first-order logic neural models like Neural Logic Machines \cite{asai22}. 
 
 Overall, the aforementioned approaches show to be competitive with baseline sequential models that use planning heuristics or state-of-the-art implementations of classical planners in terms of coverage (number of solved problems). More interestingly, some works report the inability of NN-based heuristics to outperform classical planning engines in transport-like domains wherein a tight coupling between the different objects of a planning task exists \cite{rivlin20,bonet22a,asai22}.

Frameworks that combine AI planning and RL establish a one-to-one correspondence between operators in planning and actions in RL. It is observed, as stated before, that low performance usually arises in domains with a high number of agents that need to collaborate with each other to reach the goal, also called tightly-coupled domains \cite{onaindia14}, for example \textit{logistics} or \textit{depots}.
 
Specifically, speaking about \textit{logistics}, in \cite{bonet22a} authors prove that it is not possible to achieve satisfactory results using a GNN architecture. The authors identify transitive relations without which the locality of the objects is lost. For example, when a package is inside a truck and the truck is in a city, the information about the city where the package is found fails to propagate in a transitive manner, i.e., it fails to infer that the package is indeed in the city. Results within this perspective are improved by hard-coding extra predicates (derived atoms) and including them to guarantee the transitive relations. In the mentioned example, a new atom is added to denote that a package is in a city.
 
 Interestingly, we found that in tightly-coupled domains such as \textit{logistics}, \textit{depots} or \textit{multi-blocksworld} test results from our perspective are very satisfactory when including meta-operators. In fact, if we compare the same IPC instances of the \textit{logistics} domain from \cite{bonet22a} with our approach we observe that, while they have a coverage of 17 (28) without using the hard-coded atoms, we obtain 26 (28). It should be emphasized that our process is automatic, i.e., it is not necessary to manually enrich any problem or domain to reach the goal.

It is interesting to think about why adding meta-operators actually improves the training process when it comes to using RL for planning. It is not only the fact that it allows several atomic operators to coexist at the same time, decentralizing as already mentioned in other sections the purely sequential nature of RL, but also the fact that when adding the extra operators the algorithm converges faster and yields better results.

We hypothesize that the inclusion of meta-operators opens a way of (virtual) communication between agents. As stated in \cite{rivlin20}, problems in tightly-coupled domains arise when certain resources need to be used by several agents at the same time. Since a policy is nothing more than the ``brain'' of the planning agent, when several entities can be executed in parallel (e.g. moving different vehicles) or change their state (location, mode, etc.) at the same time, the sequential nature of the RL may interfere with this notion of collaboration, thus obstructing the convergence of RL methods for this type of problems.

By introducing parallel actions we allow several entities to change their state at the same time, so that they can ``prepare'' for what is next to come. A clear example is in the \textit{logistics} domain: if we have a package being transported in a truck, which must then be picked up by another entity that transports it elsewhere, in a sequential nature the truck would leave the package at one location, and then the other entity would have to pick it up and take it away. However, how can one distinguish, given that the only known information is that the package is at one location, that the package is to be picked up by the next entity and not by the same entity? By introducing parallelism, the second entity could go to the exchange point  \textit{before} the package actually reaches that exchange point. We believe that our approach provides some sense of ``entity'' to the independent objects of a planning problem and improves agent collaboration.

On the other hand, we also wanted to hypothesize about why meta-operators also favor exploration in the state space of planning. If we think about state graphs as in \cite{geffnergraph22}, which are graphs that outline the whole structure of actions and states that the agent can take at each moment, we can consider adding meta-operators to the RL action space as introducing extra edges to these state graphs. 

In the end, training with RL is done on a trial-and-error basis, choosing actions randomly at first and seeing if it brings some benefit. By including meta-operators we allow our planning agent to do in one sampling step what previously had to be done in more (as long as the actions can coexist in parallel, obviously). This helps the convergence and thus the generalization of our methods.

To illustrate this last affirmation, we want to give another example. Let's assume an initial training iteration with a policy that is initialized with a uniform probability $p$ to each RL action. If the agent decides to apply a meta-operator of degree $L$, we arrive to a state $s$ with a probability of $p$, but in a sequential environment we would have to arrive to $s$ (following the same path) with $L$ different transitions, each one corresponding to the atomic operators that conform the meta-operator, with a $p^L$ probability. This summarizes why we think it is interesting to apply meta-operators.

\section{Conclusion and Future work}

In conclusion, we defined the notion of meta-operator as a way to enrich the RL structure that has hegemonically been used to attack the planning problem. Furthermore, we saw that it improves results from other sequentially trained models, both with randomly generated problems and with problems from the International Planning Competition. Finally, we wanted to highlight the discussion that may arise from this paper, with special emphasis on the fact that RL structures can be enriched to better match the true nature of planning.

Finally, we are aware that the inclusion of meta-operators makes the action space larger, so a balance between parallelism and size must also be found so that training times are not excessively long. For future work we would like to explore different ways of attacking RL problems with much larger, even continuous, action spaces in order to computationally be able to use meta-operators of any degree. We would also like to explore other different ways of representing planning states and how this may influence what was analyzed and to try to explain how does tuning the meta-operators reward affect in several different domains.

\section{Acknowledgments}

This work is supported by the Spanish AEI PID2021-127647NB-C22 project and Ángel Aso-Mollar is partially supported by the FPU21/04273.

\bibliography{bibliography}

\begin{thebibliography}{20}
\providecommand{\natexlab}[1]{#1}

\bibitem[{Blum and Furst(1997)}]{BlumF97}
Blum, A.; and Furst, M.~L. 1997.
\newblock {Fast Planning Through Planning Graph Analysis}.
\newblock \emph{Artif. Intell.}, 90(1-2): 281--300.

\bibitem[{Bonet and Geffner(2022)}]{geffnergraph22}
Bonet, B.; and Geffner, H. 2022.
\newblock Language-Based Causal Representation Learning.

\bibitem[{Celorrio, Aguas, and Jonsson(2019)}]{celorrio19}
Celorrio, S.~J.; Aguas, J.~S.; and Jonsson, A. 2019.
\newblock A review of generalized planning.
\newblock \emph{Knowl. Eng. Rev.}, 34: e5.

\bibitem[{Dong et~al.(2019)Dong, Mao, Lin, Wang, Li, and Zhou}]{nlmpaper}
Dong, H.; Mao, J.; Lin, T.; Wang, C.; Li, L.; and Zhou, D. 2019.
\newblock Neural Logic Machines.
\newblock \emph{CoRR}, abs/1904.11694.

\bibitem[{Eysenbach, Salakhutdinov, and Levine(2019)}]{EysenbachSL19}
Eysenbach, B.; Salakhutdinov, R.; and Levine, S. 2019.
\newblock {Search on the Replay Buffer: Bridging Planning and Reinforcement
  Learning}.
\newblock In \emph{Advances in Neural Information Processing Systems 32: Annual
  Conference on Neural Information Processing Systems 2019, NeurIPS 2019,
  December 8-14, 2019, Vancouver, BC, Canada}, 15220--15231.

\bibitem[{Franc{\`{e}}s et~al.(2019)Franc{\`{e}}s, Corr{\^{e}}a, Geissmann, and
  Pommerening}]{frances19}
Franc{\`{e}}s, G.; Corr{\^{e}}a, A.~B.; Geissmann, C.; and Pommerening, F.
  2019.
\newblock Generalized Potential Heuristics for Classical Planning.
\newblock In Kraus, S., ed., \emph{Proceedings of the Twenty-Eighth
  International Joint Conference on Artificial Intelligence, {IJCAI} 2019,
  Macao, China, August 10-16, 2019}, 5554--5561. ijcai.org.

\bibitem[{Gehring et~al.(2022)Gehring, Asai, Chitnis, Silver, Kaelbling,
  Sohrabi, and Katz}]{asai22}
Gehring, C.; Asai, M.; Chitnis, R.; Silver, T.; Kaelbling, L.~P.; Sohrabi, S.;
  and Katz, M. 2022.
\newblock Reinforcement Learning for Classical Planning: Viewing Heuristics as
  Dense Reward Generators.
\newblock In Kumar, A.; Thi{\'{e}}baux, S.; Varakantham, P.; and Yeoh, W.,
  eds., \emph{Proceedings of the Thirty-Second International Conference on
  Automated Planning and Scheduling, {ICAPS} 2022, Singapore (virtual), June
  13-24, 2022}, 588--596. {AAAI} Press.

\bibitem[{Icarte et~al.(2022)Icarte, Klassen, Valenzano, and
  McIlraith}]{rewardmachines}
Icarte, R.~T.; Klassen, T.~Q.; Valenzano, R.~A.; and McIlraith, S.~A. 2022.
\newblock Reward Machines: Exploiting Reward Function Structure in
  Reinforcement Learning.
\newblock \emph{J. Artif. Intell. Res.}, 73: 173--208.

\bibitem[{Lee et~al.(2022)Lee, Katz, Agravante, Liu, Tasse, Klinger, and
  Sohrabi}]{hierarchicalRL}
Lee, J.; Katz, M.; Agravante, D.~J.; Liu, M.; Tasse, G.~N.; Klinger, T.; and
  Sohrabi, S. 2022.
\newblock Hierarchical Reinforcement Learning with AI Planning Models.

\bibitem[{Moerland et~al.(2023)Moerland, Broekens, Plaat, and
  Jonker}]{modelbasedrlsurvey}
Moerland, T.~M.; Broekens, J.; Plaat, A.; and Jonker, C.~M. 2023.
\newblock Model-based Reinforcement Learning: {A} Survey.
\newblock \emph{Found. Trends Mach. Learn.}, 16(1): 1--118.

\bibitem[{Ng and Petrick(2019)}]{alvin19}
Ng, J. H.~A.; and Petrick, R. P.~A. 2019.
\newblock Incremental Learning of Planning Actions in Model-Based Reinforcement
  Learning.
\newblock In Kraus, S., ed., \emph{Proceedings of the Twenty-Eighth
  International Joint Conference on Artificial Intelligence, {IJCAI} 2019,
  Macao, China, August 10-16, 2019}, 3195--3201. ijcai.org.

\bibitem[{Puterman(1990)}]{mdpsbasic}
Puterman, M.~L. 1990.
\newblock Chapter 8 Markov decision processes.
\newblock In \emph{Stochastic Models}, volume~2 of \emph{Handbooks in
  Operations Research and Management Science}, 331--434. Elsevier.

\bibitem[{Rivlin, Hazan, and Karpas(2020)}]{rivlin20}
Rivlin, O.; Hazan, T.; and Karpas, E. 2020.
\newblock Generalized Planning With Deep Reinforcement Learning.
\newblock \emph{CoRR}, abs/2005.02305.

\bibitem[{Seipp, Torralba, and Hoffmann(2022)}]{pddlgenerators}
Seipp, J.; Torralba, {\'A}.; and Hoffmann, J. 2022.
\newblock {PDDL} Generators.
\newblock \url{https://doi.org/10.5281/zenodo.6382173}.

\bibitem[{Silver and Chitnis(2020)}]{pddlgym}
Silver, T.; and Chitnis, R. 2020.
\newblock PDDLGym: Gym Environments from {PDDL} Problems.
\newblock \emph{CoRR}, abs/2002.06432.

\bibitem[{St{\aa}hlberg, Bonet, and Geffner(2022{\natexlab{a}})}]{bonet22b}
St{\aa}hlberg, S.; Bonet, B.; and Geffner, H. 2022{\natexlab{a}}.
\newblock Learning General Optimal Policies with Graph Neural Networks:
  Expressive Power, Transparency, and Limits.
\newblock In Kumar, A.; Thi{\'{e}}baux, S.; Varakantham, P.; and Yeoh, W.,
  eds., \emph{Proceedings of the Thirty-Second International Conference on
  Automated Planning and Scheduling, {ICAPS} 2022, Singapore (virtual), June
  13-24, 2022}, 629--637. {AAAI} Press.

\bibitem[{St{\aa}hlberg, Bonet, and Geffner(2022{\natexlab{b}})}]{bonet22a}
St{\aa}hlberg, S.; Bonet, B.; and Geffner, H. 2022{\natexlab{b}}.
\newblock Learning Generalized Policies without Supervision Using GNNs.
\newblock In Kern{-}Isberner, G.; Lakemeyer, G.; and Meyer, T., eds.,
  \emph{Proceedings of the 19th International Conference on Principles of
  Knowledge Representation and Reasoning, {KR} 2022, Haifa, Israel. July 31 -
  August 5, 2022}.

\bibitem[{Sutton and Barto(1998)}]{sutton98}
Sutton, R.~S.; and Barto, A.~G. 1998.
\newblock \emph{Reinforcement learning - an introduction}.
\newblock Adaptive computation and machine learning. {MIT} Press.
\newblock ISBN 978-0-262-19398-6.

\bibitem[{Torre{\~{n}}o, Onaindia, and Sapena(2014)}]{onaindia14}
Torre{\~{n}}o, A.; Onaindia, E.; and Sapena, {\'{O}}. 2014.
\newblock {FMAP}: Distributed cooperative multi-agent planning.
\newblock \emph{Applied Intelligence}, 41(2): 606--626.

\bibitem[{Zhou et~al.(2020)Zhou, Cui, Hu, Zhang, Yang, Liu, Wang, Li, and
  Sun}]{gnnsurvey}
Zhou, J.; Cui, G.; Hu, S.; Zhang, Z.; Yang, C.; Liu, Z.; Wang, L.; Li, C.; and
  Sun, M. 2020.
\newblock Graph neural networks: {A} review of methods and applications.
\newblock \emph{{AI} Open}, 1: 57--81.

\end{thebibliography}

\end{document}